\documentclass[runningheads]{llncs}
\usepackage{graphicx}
\usepackage{amsmath}
\usepackage{amssymb}
\usepackage{booktabs}
\usepackage{listings}
\usepackage{dsfont}

\usepackage{mathabx}

\usepackage{xspace}

\usepackage{booktabs}
\begin{document}
%
\title{Lightweight Transformer Backbone for Medical Object Detection}
%
%
%


%

\author{Yifan Zhang\inst{1,2}{$^*$}{$^\dag$} \and
Haoyu Dong\inst{1}$^*$ \and
Nicholas Konz\inst{3} \and
Hanxue Gu\inst{3} \and \\
Maciej A. Mazurowski\inst{1,3,4,5}}
\authorrunning{Y. Zhang et al.}
%
\institute{Department of Radiology, Duke University, USA \and
Department of Computer Science, Vanderbilt University, USA \and
Department of Electrical and Computer Engineering, Duke University, USA \and
Department of Biostatistics \& Bioinformatics, Duke University, USA \and
Department of Computer Science, Duke University, USA
\email{yifan.zhang.2@vanderbilt.edu,hd108@duke.edu} \email{\{nicholas.konz,hg119,maciej.mazurowski\}@duke.edu}}

\def\thefootnote{*}\footnotetext{Equal contribution}\def\thefootnote{\arabic{footnote}}
\def\thefootnote{\dag}\footnotetext{Corresponding author}\def\thefootnote{\arabic{footnote}}

 \maketitle              

\begin{abstract}

Lesion detection in digital breast tomosynthesis (DBT) is an important and a challenging problem characterized by a low prevalence of images containing tumors. Due to the label scarcity problem, large deep learning models and computationally intensive algorithms are likely to fail when applied to this task. In this paper, we present a practical yet lightweight backbone to improve the accuracy of tumor detection. Specifically, we propose a novel modification of visual transformer (ViT) on image feature patches to connect the feature patches of a tumor with healthy backgrounds of breast images and form a more robust backbone for tumor detection. To the best of our knowledge, our model is the first work of Transformer backbone object detection for medical imaging. Our experiments show that this model can considerably improve the accuracy of lesion detection and reduce the amount of labeled data required in typical ViT. We further show that with additional augmented tumor data, our model significantly outperforms the Faster R-CNN model and state-of-the-art SWIN transformer model.

\end{abstract}

\section{Introduction}

Medical Object Detection (OD) distinguish the object of interest from medical images, which is important in the downstream medical applications such as diagnosis. In the era of big data, hospitals are able to gather a large amount of images to train detection models and use them to assist radiologist with different disciplines \cite{yang2021artificial}. However, training such models usually needs accurate tumor bounding boxes, which is labor-intensive and resource consuming to annotate. The scarcity of available bounding boxes constrains the volume of models and limits the overall model performance, eliciting the need for an innovative structure to perform effective medical object detection with the most efficient model design.

Having a reliable model architecture for medical OD is essential. The classic methods for object detection utilize Convolutional Neural Networks (CNN) \cite{albawi2017understanding,cai2018cascade,chen2018domain} to select a considerable number of regions for location prediction. To reduce the number of assigned areas, Region-Based Convolutional Neural Networks (R-CNN)  \cite{girshick2014rich} choose an exact number of proposed regions into CNN to improve the model efficiency. To further solve the drawbacks of computational efficiency in R-CNN, Fast R-CNN \cite{girshick2015fast} and Faster R-CNN \cite{ren2015faster} were proposed to feed the input image directly into CNN to generate a convolutional feature map for proposal regions. Looking at the complete picture and predicting a class probability in each grid, YOLO \cite{redmon2016you} outperforms the previous methods and becomes the most efficient and effective framework in CNN-based detection.

Recent studies in Transformer \cite{vaswani2017attention}, an attention based neural network structure, have advanced the performance in OD. 
With the evolution of Transformer on vision, Visual Transformer (ViT) \cite{dosovitskiy2020image} is proposed to model long-term dependencies of image patches. The latest state-of-the-art (SOTA) Transformer method on vision, called SWIN Transformer \cite{liu2021swin}, uses shifted windows to construct hierarchical visual representations for downstream application and is beneficial for most modeling in natural images. However, ViT and SWIN are constrained by a considerable amount of required data, which is usually unreachable in medical object detection scenarios. For example, both methods targeted on COCO \cite{Lin2014MicrosoftCC}, which consists of 118K labelled training samples, while most medical dataset consists of less than 1K labelled samples.

To improve the performance of CNN-based medical OD models and address the data hunger in the attention mechanism, we introduce a lightweight Transformer backbone for improving detection accuracy without extra annotations. Specifically, we replace the feature pyramid network (FPN) module by using a weighted sum strategy to integrate the features from different layers instead of summing them equally. To achieve this goal, we use feature rearrangement and reconstruction to reshape and restore the feature maps of ResNet into and from feature patches, which unifies the model on outputs of each ResNet layer with only one ViT layer. The reconstruction task also allows the model to fully utilize the training data. Besides, we introduce novel lightweight attention of ViT to enhance the representations of rearranged feature patches. The experiments demonstrate that using our model for medical object detection is highly promising, achieving significantly better performance to the Faster R-CNN method. It also considerably outperforms the SOTA ViT model (SWIN).


\section{Methodology}

In this paper, we intend to provide a lightweight Transformer backbone for medical object detection with limited positively labelled data. This section presents the key components of the proposed method.

\subsection{Overview of Proposed Method}



\begin{figure*}
\centering
\includegraphics[width=1\textwidth]{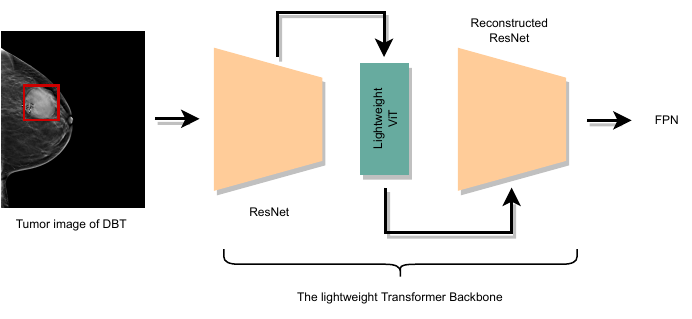}
\caption{Illustration of our lightweight Transformer backbone. The ResNet and reconstructed ResNet have the same shape. The black arrows represent information flow.}\label{fig1}
\end{figure*}


As shown in Fig.\ref{fig1}, we propose a new lightweight ViT backbone for coupling with the feature pyramid network in order to improve detection performance. In our pipeline,  raw images are fed into a ResNet \cite{he2016deep} to generate feature maps corresponding to the activation map of each hidden layer. Because spatial attention can significantly improve the connections between pixels, before feeding these out-puts into the next FPN \cite{lin2017feature} that uses multi-scale pyramidal hierarchy to construct feature pyramids for Region Proposal Network (RPN) and Region of Interest (RoI) pooling \cite{ren2015faster}, we apply attention on image feature patches of the outputs of ResNet to improve the hidden representations of each inputs to FPN.

\subsection{Feature map rearrangement \& reconstruction}

\begin{figure*}
\centering
\includegraphics[width=0.9\textwidth]{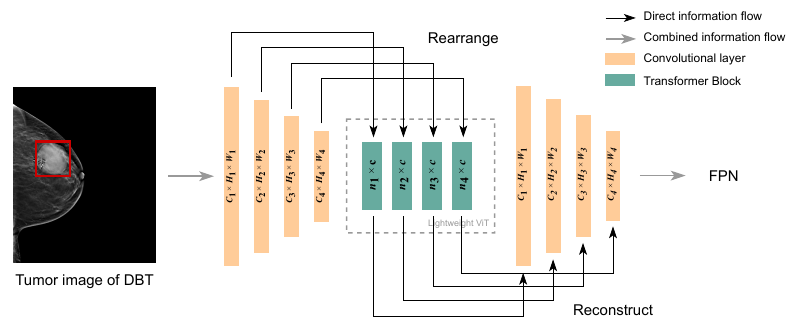}
\caption{Feature map rearrangement and reconstruction module. Both Rearrangement and reconstruction connect to the same lightweight ViT module. We omit the batch size for clarity.}\label{fig2}
\end{figure*}

In this section, we introduce how the feature maps of ResNet are rearranged as feature patches and fit the inputs of our ViT, and how the feature patches are reconstructed with the original shape of feature maps generated by ResNet. The feature rearrangement and reconstruction process of our model are shown in Fig.\ref{fig2}.

\textbf{Feature map rearrangement} The ResNet outputs have the shapes of $(B, C_k, H_k, W_k)$, in which $B$ is the batch size, $C_k$, $H_k$ and $W_k$ denote the number of channels, feature map height and width in the $k^{th}$ layer. Each patch of feature map in the ResNet outputs will be subjected to the following rearrangement in the first section:

\begin{equation}
\boldsymbol{z_k} = [\boldsymbol{x}_{kp}^1\boldsymbol{E};  \boldsymbol{x}_{kp}^2\boldsymbol{E}; ... ; \boldsymbol{x}_{kp}^{N_k} \boldsymbol{E}],
\end{equation}

where $\boldsymbol{x_{kp}}$ and $\boldsymbol{z_k}$ stand for patches of feature map $\boldsymbol{x}$ and feature patches $\boldsymbol{z}$ in the $k^{th}$ layer. $\boldsymbol{E}$ denotes a \emph{feature map rearrangement embedding}, $\boldsymbol{E}\in\mathbb{R}^{(w \cdot h \cdot C_k) \times c}$, in which $w$ and $h$ are width and height of a single patch.

The feature map transformation embedding is for transforming an original feature map to a feature map on patches. To be more specific, it rearranges the shape of feature map by the formula below: 

\begin{equation}
(B, C_k, H_k, W_k) \rightarrow (B, n_k, c).
\end{equation}

Here $n_k$ is the number of feature patches in $k^{th}$ layer, and $C_k$ is the number of channels in the current hidden layer, which is computed as

\begin{equation}
C_k = c \cdot 2^{(k-1)}, c = 256.
\end{equation}

Since the output of ResNet follows a pyramid structure, with the increment in hidden dimension and stride over spatial dimension, we use divisible numbers of the spatial size of the last-layer representation as the patch size, which is (5, 4). Then, all shallow layers are first resized along hidden dimension to match the depth information, i.e., $(256\times2, n) \rightarrow (256, (2 \times n))$. Therefore, the number of feature patches for the $k^{th}$ layer will be

\begin{equation}
n_k = (H_k / h) \cdot (W_k / w) \cdot 2^{(k-1)}, h = 5, w = 4.
\end{equation}

\textbf{Feature map reconstruction} After the rearrangement and the lightweight Transformer module, feature patches will be reconstructed to their original shape as shown in the formula   

\begin{equation}
(B, n_k, c) \rightarrow (B, C_k, H_k, W_k).
\end{equation}

Because the only quantity that changes across layers is the total number of feature patches, this design ensures weight sharing of the ViT module.

\subsection{Lightweight Transformer on feature patches}

\begin{figure*}
\centering
\includegraphics[width=1\textwidth]{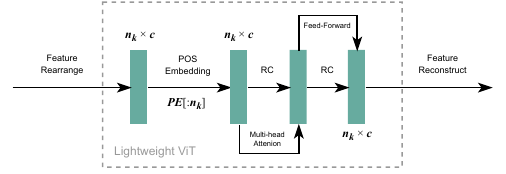}
\caption{The lightweight Transformer module. It consists of positional embedding, attention and feed-forward. All vectors omit the batch size dimension for clarity.}\label{fig3}
\end{figure*}

We design a lightweight Transformer module to enhance the representation of feature patches, consisting of positional embedding, attention, and feedforward components. The lightweight ViT module in our model is illustrated in Fig.\ref{fig3}. 

\textbf{Positional Embedding} After the feature rearrangement, feature patches has the shape of $(B, n_k, c)$. As with BERT \cite{devlin2018bert} and ViT \cite{dosovitskiy2020image}, we append a learnable positional embedding $E_{pos}$ to assist the network in remembering the locations of individual patches, as 

\begin{equation}
\boldsymbol{z_k} = \boldsymbol{z_k} + \boldsymbol{E}_{pos}{[:n_k]},
\end{equation}

where $\boldsymbol{E}_{pos}$ is a shared positional embedding for feature patches of each rearranged ResNet feature map. It has a maximum length of maximal number of feature patches $N \cdot 2^{min(k)-1}$, where $N$ denotes the number of patches in the first ResNet layer, which is 4096, and $min(k) = 1$.

\textbf{Multi-head Attention} We adopt the multi-head self-attention mechanism in ViT to jointly infer attention from different representation subspaces. The output of the self-attention is a scaled dot-product:

\begin{equation}
Attention(Q, K, V) = softmax(\frac{QK^T}{\sqrt{d_k}})V,
\end{equation}

where $Q, K, V \in \mathbb{R}^{(w \cdot h \cdot c)}$ are query, key and value embeddings, and $\sqrt{d_k}$ is the dimension of the key vector $k$ and query vector $q$. We extend it to the multi-head attention:

\begin{equation}
    MultiHead(Q, K, V) = Concat(head_1, ..., head_I)W^O,
\end{equation}

where

\begin{equation}
    head_i = Attention(Q W^Q_i, K W^K_i, V W^V_i),
\end{equation}

Here $W^Q_i, W^K_i, W^V_i, W^O$ denote trainable parameters corresponding to $Q, K, V$ in the $i_{th}$ attention head, and the output. In this study, we use $I=8$. The results from multiple heads are concatenated and then transformed with a feed-forward network.

\textbf{Feed-forward} We use the same input and output dimension in the feed-forward layer to keep the original shape of feature patches. The feed-forward layer adopts one dimension reduction layer to project the patch dimension into a lower dimension space. When the fixed number of channels $c = 256$, we have patch dimension as 5120, so we define hidden dimension as 2560. We use GeLU as non-linear activation and a dropout layer to increase the generalizability. Following those layers, we use another dimension-raising layer to restore the original dimension of feature patches.

For both the multi-head attention and feed-forward layers, we adopt the layer pre-norm and residual connection (RC) in ViT. We observe that using the lightweight Transformer on feature patches, the output representation of each layer is significantly enriched by the combination of tumor information and different locations in medical images.

\section{Experiments and Results}


\subsection{Dataset and Evaluation Metrics}

\begin{figure*}
\centering
\includegraphics[width=0.8\textwidth]{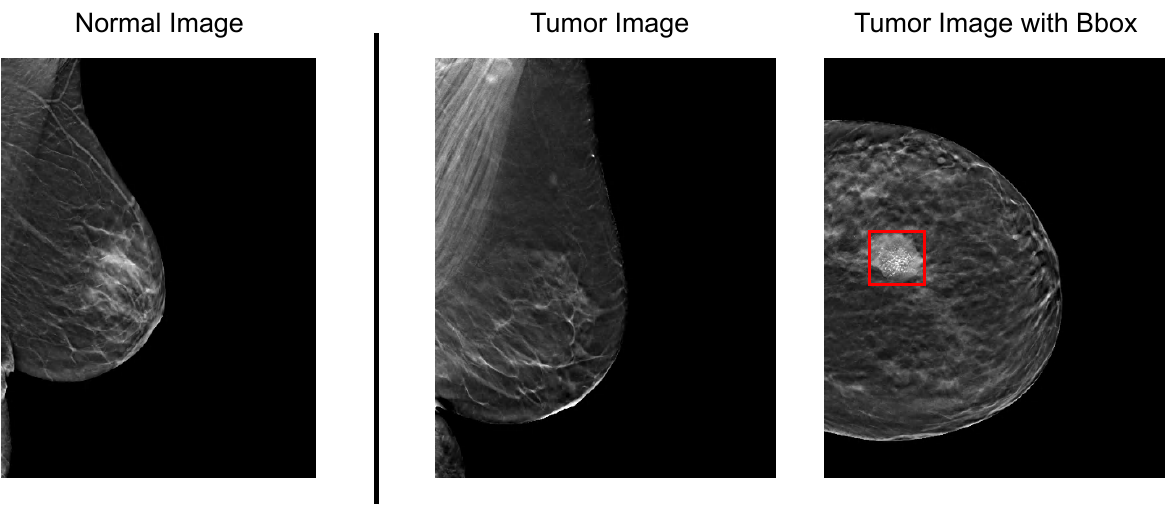}
\caption{The examples of normal and tumor images from BCS-DBT. Res box denotes the bounding box of the tumor image.}\label{fig4}
\end{figure*}

We use a publicly available dataset of breast cancer screening scans: the Digital Breast Tomosynthesis (BCS-DBT) dataset \cite{buda2020data}. The BCS-DBT dataset comprises cancer cases that are normal, actionable, non biopsy-proven, and biopsy-proven. It contains 22032 breast tomosynthesis scans from 5060 individuals, with each scan containing up to 4 anatomical views and dozens of spatially-aligned slices in each view. Fig \ref{fig4} shows three examples of DBT images.

In our study, we use only the tumor slices with bounding boxes. In BCS-DBT, there are 299 tumor slices with 346 bounding boxes. As indicated in \cite{buda2020data}, we split the data into a training set with 233 tumor slices and 274 bounding boxes, and a validation set with 75 tumor slices and 75 bounding boxes.

We use the AP (Average Precision) metrics for object detection as a quantitative study of our model. There are metrics relying on IoU (Intersection over Union) that describes the intersection of ground-truth bounding boxes and predicted bounding boxes from models. The IoU formula is known as
\begin{equation}
IoU(A,B) = \frac{(A \cap B)}{(A \cup B)},
\end{equation}
where A and B are ground-truth bounding boxes and preidcted bounding boxes from models, respectively. $IoU(A,B) \in [0, 1]$.

In our study, we employ AP, AP50, AP75, APm, and APl as analysis criteria, with AP50 serving as an indicator of model performance. Here, AP50 and AP75 counts the samples that have at least 0.5 and 0.75 IoU areas respectively and AP is the average of AP50 to AP90 with step size 5. APm and APl are for medium objects with areas $ \in [32^2, 96^2]$ and large objects with areas more than $96^2$.

\subsection{Implementation Details}

We use Detectron2 \cite{wu2019detectron2} as our object detection framework. In our study, we implement two comparative models to demonstrate the competitiveness and generalization ability of our model. We use the other official implementations and default hyperparameters for training all three models in the Detectron2.

\textbf{Faster R-CNN} We adopt the Faster R-CNN in Detectron2 using CNN as the backbone feature extractor. It utilizes a pretrained ResNet50 to get feature maps. During training, we set the batch size of 4 for the best performance. 

\textbf{SWIN Transformer} We implement a pretrained SWIN Transformer as the feature extractor of a Faster R-CNN model. It utilizes a windows size of 7 and an embedding dimension of 96. During training, we set the batch size of 4 and the learning rate of 0.001 for the best performance.

The original Detectron2 framework has random data augmentation that resizes images to 8 various shapes, some of which may be incompatible with our preset patch size. The widths and heights of hidden feature maps should be an integer multiple of the width and heights of the preset patch. As a result, we remove the random data resize in the experiments.

\subsection{Experimental Results}

We compare our proposed lightweight Transformer backbone to Faster R-CNN and SWIN Transformer in tumor detection and evaluate the performance of the models. We further prove the effectiveness of our model with additional augmented tumor data. The detailed method for tumor augmentation is attached in the appendix for clarity.

\textbf{Comparative Studies} Tab. \ref{tab1} summarizes the qualitative results of all four methods in tumor detection performance. The AP50 shows that our lightweight Transformer approach has achieved significantly more accurate detection of tumors, improving 7.2\% (+2.49) on Faster R-CNN. The standard deviations show that our Lightweight Transformer method has more stability in strict criterion AP75 and medium object detection APm, whereas relatively diverse in other criteria, including AP, AP50, and APl.

On the contrary, the SWIN Transformer backbone performs considerably worse than both Faster R-CNN and Lightweight Transformer backbones. The AP50 shows that the performance of SWIN Transformer backbone drops by 25.26 and 27.75 comparing with Faster R-CNN and Lightweight Transformer backbones respectively, indicating the ineffectiveness of direct application of Transformer image feature extraction due to the scarity of available data. 

\begin{table*}[t]
\caption{The quantitative metrics of our lightweight backbone and other methods (mean $\pm$ std). Each simulation was performed 5 times for computing the means and standard deviation of criteria.}\label{tab1}
\centering
\setlength{\abovecaptionskip}{0pt}   
\setlength{\belowcaptionskip}{10pt}
\begin{tabular}{c|c|c|c|c|c}
\toprule
Method & AP & AP50 & AP75 & APm & APl  \\\hline \hline
Faster R-CNN & 11.78 \tiny($\pm$ 1.08) & 39.55 \tiny($\pm$ 1.15) & 4.06 \tiny($\pm$ 2.24) & \textbf{9.36} \tiny($\pm$ 3.12) & 12.12 \tiny($\pm$ 1.03) \\
SWIN Transformer &  4.28 \tiny($\pm$ 0.85) & 14.29 \tiny($\pm$ 1.38) & 2.01 \tiny($\pm$ 0.97)  & 1.44 \tiny($\pm$ 0.96) & 4.71 \tiny($\pm$ 0.75) \\
Lightweight (Ours) & \textbf{13.71} \tiny($\pm$ 1.20) & \textbf{42.04} \tiny($\pm$ 2.74) & \textbf{4.73} \tiny($\pm$ 1.66) & 6.20 \tiny($\pm$  3.07) & \textbf{14.45} \tiny($\pm$ 1.62) \\
\bottomrule
\end{tabular}
\end{table*}

\begin{table*}[h]
\caption{The main metrics and their difference of our lightweight backbone and other methods on augmented tumor dataset (mean $\pm$ std). Each simulation was performed 5 times for computing the means and standard deviation of criteria.}\label{tab2}
\centering
\setlength{\abovecaptionskip}{0pt}   
\setlength{\belowcaptionskip}{10pt}
\begin{tabular}{c|c|c|c|c}
\toprule
Method & AP & AP50 & AP Change & AP50 Change \\\hline \hline
Faster R-CNN & 12.84 \tiny($\pm$ 2.36) & 41.42 \tiny($\pm$ 1.96) & +1.06 & + 1.87\\
SWIN Transformer & 4.18 \tiny($\pm$ 1.08) & 14.22 \tiny($\pm$ 1.90) & - 0.1  & - 0.07 \\
Lightweight (Ours) & \textbf{13.41} \tiny($\pm$ 0.82) & \textbf{44.03} \tiny($\pm$ 2.12) & - 0.3  & + 1.99\\
\bottomrule
\end{tabular}
\end{table*}

\textbf{Evaluation on Augmented Dataset} We further compare our method with Faster R-CNN and SWIN using a 4x larger augmented training dataset through inserting tumors into normal images.  The details of insertion can be found in the appendix. This leads to  932 tumor slices and 973 bounding boxes for training, while the validation set is kept the same. Tab. \ref{tab2} shows that with additional labeled data, the lightweight Transformer method performs better than Faster R-CNN baseline in both AP (13.41) and AP50 (44.03), with higher increment (+1.99) in AP50 as well.

At the same time, the performance of the SWIN Transformer doesn't change much on AP (-0.1) and AP50 (-0.07), showing that the direct application of Transformer-based models as feature extractors needs a considerably more extensive dataset.

\section{Conclusion}

We proposed the lightweight Transformer backbone in this work to improve the performance of the medical object detection model in the context of breast tumor detection. As a novel backbone for improving high-resolution breast tumor detection's performance and stability and achieving higher performance on existing backbones without extra tumor annotations, our techniques provides a new idea for applying attention to related problems. We further prove that the direct application of Transformer-based methods on medical object detection requires a larger dataset, demonstrating the advantages of our proposed method.

%
%
%





\bibliographystyle{splncs04}
\bibliography{egbib}

\end{document}